\pdfoutput=1

\documentclass[11pt]{article}
\PassOptionsToPackage{table,dvipsnames}{xcolor}
\usepackage{xcolor}

\usepackage{CJKutf8}

\usepackage[preprint]{acl} 

\usepackage{times}
\usepackage{latexsym}
\usepackage{longtable}
\usepackage[T1]{fontenc}

\usepackage[utf8]{inputenc}

\usepackage{microtype}

\usepackage{inconsolata}

\usepackage{graphicx}

\usepackage{amsmath}
\usepackage{amssymb}

\usepackage{booktabs}

\definecolor{MorandiRed}{RGB}{215,157,164}
\definecolor{MorandiYellow}{RGB}{198,175,117}
\definecolor{MorandiBlue}{RGB}{118,134,146}
\definecolor{MorandiGreen}{RGB}{158,203,176}

\usepackage{colortbl}
\usepackage{array}
\usepackage{multirow}

\usepackage{soul} 
\usepackage{pifont}

\title{LegalChainReasoner: 
A Legal Chain-guided Framework for Criminal Judicial Opinion Generation}

\author{
 \textbf{Weizhe Shi\textsuperscript{1}}\thanks{Equal contribution.},
 \textbf{Qiqi Wang\textsuperscript{2}}\footnotemark[1]\thanks{Corresponding author.},
 \textbf{Yihong Pan\textsuperscript{1}},
 \textbf{Qian Liu\textsuperscript{1}}\footnotemark[2],
 \textbf{Kaiqi Zhao\textsuperscript{3}}\footnotemark[2]
\\
 \textsuperscript{1}The University of Auckland, New Zealand
 \\
 \textsuperscript{2}Nankai University, China
 \\
 \textsuperscript{3}Harbin Institute of Technology, Shenzhen, China
\\
 \texttt{\{wshi590, ypan317\}@aucklanduni.ac.nz }
 \texttt{qiqi.wang@nankai.edu.cn }
 \\
 \texttt{liu.qian@auckland.ac.nz }
 \texttt{zhaokaiqi@hit.edu.cn}
}

\begin{document}
\begin{CJK*}{UTF8}{gbsn}
\maketitle
\begin{abstract}
A criminal judicial opinion represents the judge's disposition of a case, including the decision rationale and sentencing. Automatically generating such opinions can assist in analyzing sentencing consistency and provide judges with references to similar past cases. However, current research typically approaches this task by dividing it into two isolated subtasks: legal reasoning and sentencing prediction. This separation often leads to inconsistency between the reasoning and predictions, failing to meet real-world judicial requirements. Furthermore, prior studies rely on manually curated knowledge to enhance applicability, yet such methods remain limited in practical deployment. 
To address these limitations and better align with legal practice, we propose a new LegalAI task: Judicial Opinion Generation, which simultaneously produces both legal reasoning and sentencing decisions.
To achieve this, we introduce LegalChainReasoner, a framework that applies structured legal chains to guide the model through comprehensive case assessments. By integrating factual premises, composite legal conditions, and sentencing conclusions, our approach ensures flexible knowledge injection and end-to-end opinion generation. Experiments on two real-world and open-source Chinese legal case datasets demonstrate that our method outperforms baseline models\footnote{The code and data used in this paper will be released after paper acceptance.}. 
\end{abstract}

\section{Introduction}

In criminal judicial systems, particularly within Civil Law jurisdictions, legal judgments inherently integrate two inseparable elements: structured legal reasoning and consequent sentencing decisions~\cite{cui2023survey}. This integration reflects the fundamental nature of judicial decision-making, where sentencing conclusions must be derived from and justified by systematic application of statutory provisions to established facts.
\begin{figure}[t]
\centering
\includegraphics[width=\linewidth]{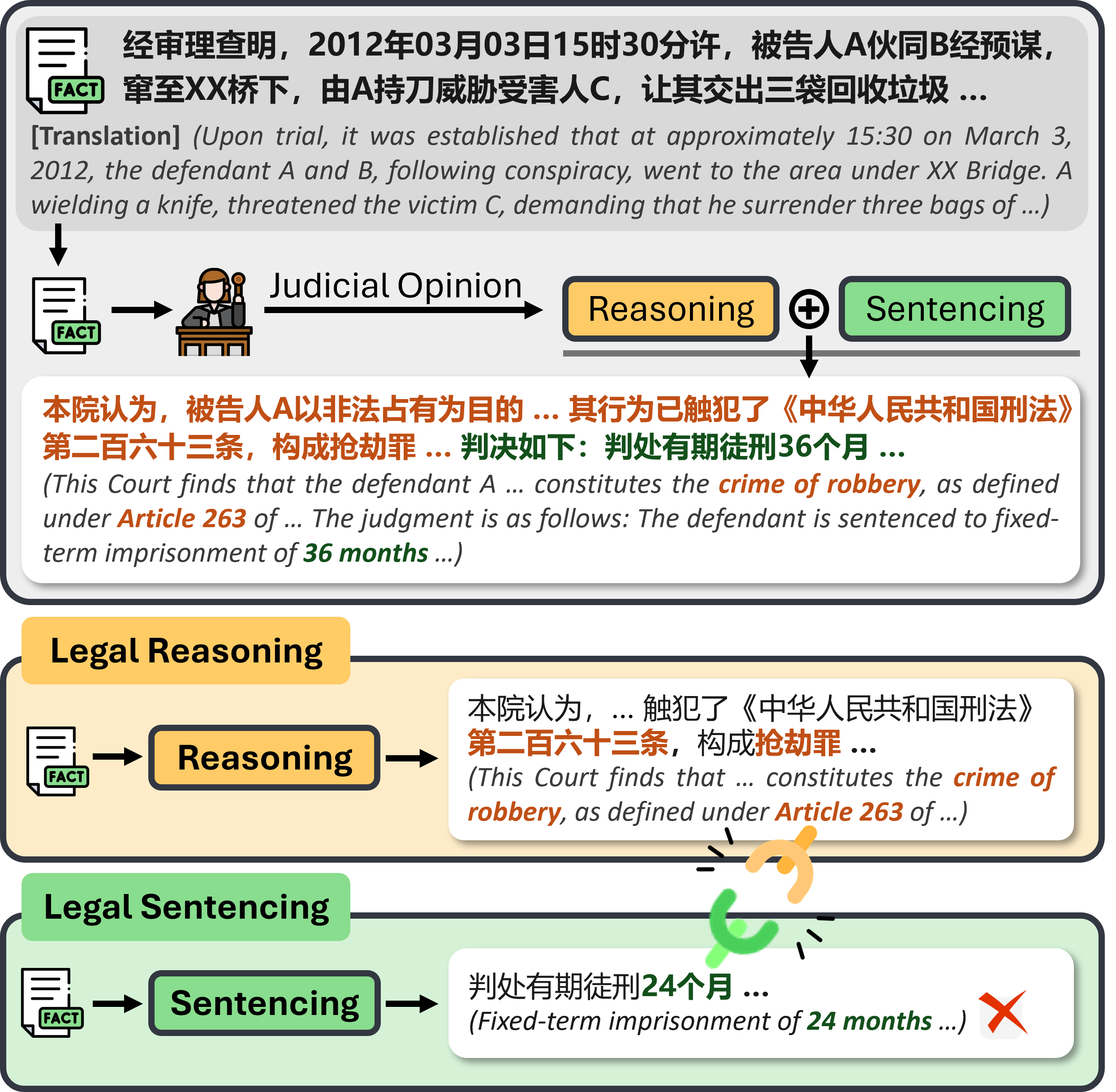}
\caption{Comparison between traditional approaches and the real-world practice. In traditional approach, legal reasoning and sentencing prediction are treated as separate tasks with independent models. In contrast, real-world judicial practice jointly consider these into a unified process, where sentencing decisions are derived from and justified by reasoning applied to case facts.}
\vspace{-1em}
\label{fig:judicial_opinion_generation}
\end{figure}
Despite this intrinsic connection, current LegalAI research artificially divides this coherent process into isolated tasks. The legal reasoning task~\cite{ye2018interpretable, wu2020biased, li2024enhancing, yue2024event} focuses exclusively on constructing plausible arguments without addressing sentencing implications, while the sentencing prediction task~\cite{liu2022augmenting, wu2023precedent, yue2021neurjudge, sun2024chinese} generates penalty terms without providing necessary legal justifications. This separation creates two critical problems: (1) reasoning models may produce analyses inconsistent with the possible sentencing outcome, and (2) prediction models generate sentences lacking transparent legal foundations. Such fragmentation fundamentally contradicts the judicial principle that sentencing must be grounded in and justified by proper legal reasoning.

Beyond the task separation, two additional limitations have also occurred in existing approaches. First, most of the legal knowledge incorporated in current methods remains manually designed and relatively simplistic~\cite{gan2021judgment}. Methods using such knowledge often rely on predefined rules or direct citations of legal provisions rather than abstracting essential reasoning patterns. This surface-level integration fails to capture the deeper logical structures underlying legal judgments, limiting models' ability to generalize across cases with subtle yet legally significant distinctions. 
Besides, current approaches lack explainability in both tasks, particularly in sentence prediction. Some studies have attempted to introduce manually defined generation processes to improve explainability in reasoning generation~\cite{xu2024divide}. Nevertheless, most existing research on sentence prediction primarily focuses on extracting semantic features from factual information, often overlooking the underlying reasoning mechanisms and the interpretability of the predicted outcomes. This limits the transparency and trustworthiness of these models, especially in the criminal domains.

To address these fundamental limitations, a novel Criminal Judicial Opinion Generation task is proposed in this paper, wherein legal reasoning and sentencing prediction are integrated into one paradigm like human judges, as illustrated in Figure~\ref{fig:judicial_opinion_generation}. This task offers two significant advantages: (1) sentencing predictions become explicitly grounded in legal reasoning, enhancing both interpretability and jurisprudential validity; and (2) the approach mirrors authentic judicial processes, where verdicts are derived through structured legal analysis rather than isolated numerical predictions.

To solve this task, we propose the LegalChainReasoner, a comprehensive framework comprising two principal components: the Structured Legal Chain and Chain-Aware encoding method. The Structured Legal Chains are developed by transforming statutory provisions into \textit{premise}-\textit{situation}-\textit{conclusion} triplets, based on the Legal Norm theory~\cite{keuth1974on, raz1980theconcept}. These chains formalize implicit legal inference patterns into explicit structured representations, enabling the model to learn step-by-step judicial decision-making processes. The Chain-Aware encoding with multi-linear transformation layers is designed to capture complex relationships between legal elements while applying crime-specific reasoning transformations, resulting in an effective jurisprudential alignment with established judicial reasoning patterns.
The main contributions of this work are:
\begin{itemize}
    \item We define a criminal judicial opinion generation task that produces legal reasoning and sentencing prediction in a single process. The task ensures the predictions are supported by judicial analyses.
    \item We reveal the importance of structured legal chains in supporting interpretable and reliable judicial opinion generation. By formalizing statutory provisions into premise-situation-conclusion triplets, we propose a chain-aware encoding method that effectively helps reasoning in the judicial decision-making process.
    \item We evaluate our proposed method on two open-source, real-world datasets and achieve improvements in both the quality of generated judicial opinions and sentencing prediction.
\end{itemize}

\section{Methodology}

\paragraph{Task Definition}
Given a set of criminal legal case factual descriptions $\mathcal{F} =\{F_1, F_2, \cdots, F_i, \cdots\}$, the Criminal Judicial Opinion Generation task is formulated as the process of generating judicial opinion $O_i$ for each fact $F_i$, by jointly considering legal reasoning and sentencing determination. The task can then be defined as the function $f: F_i \rightarrow O_i$. The judicial opinion $O_i$ includes the case reasoning and sentence prediction. 




\begin{figure}[t]
\centering
\includegraphics[width=\linewidth]{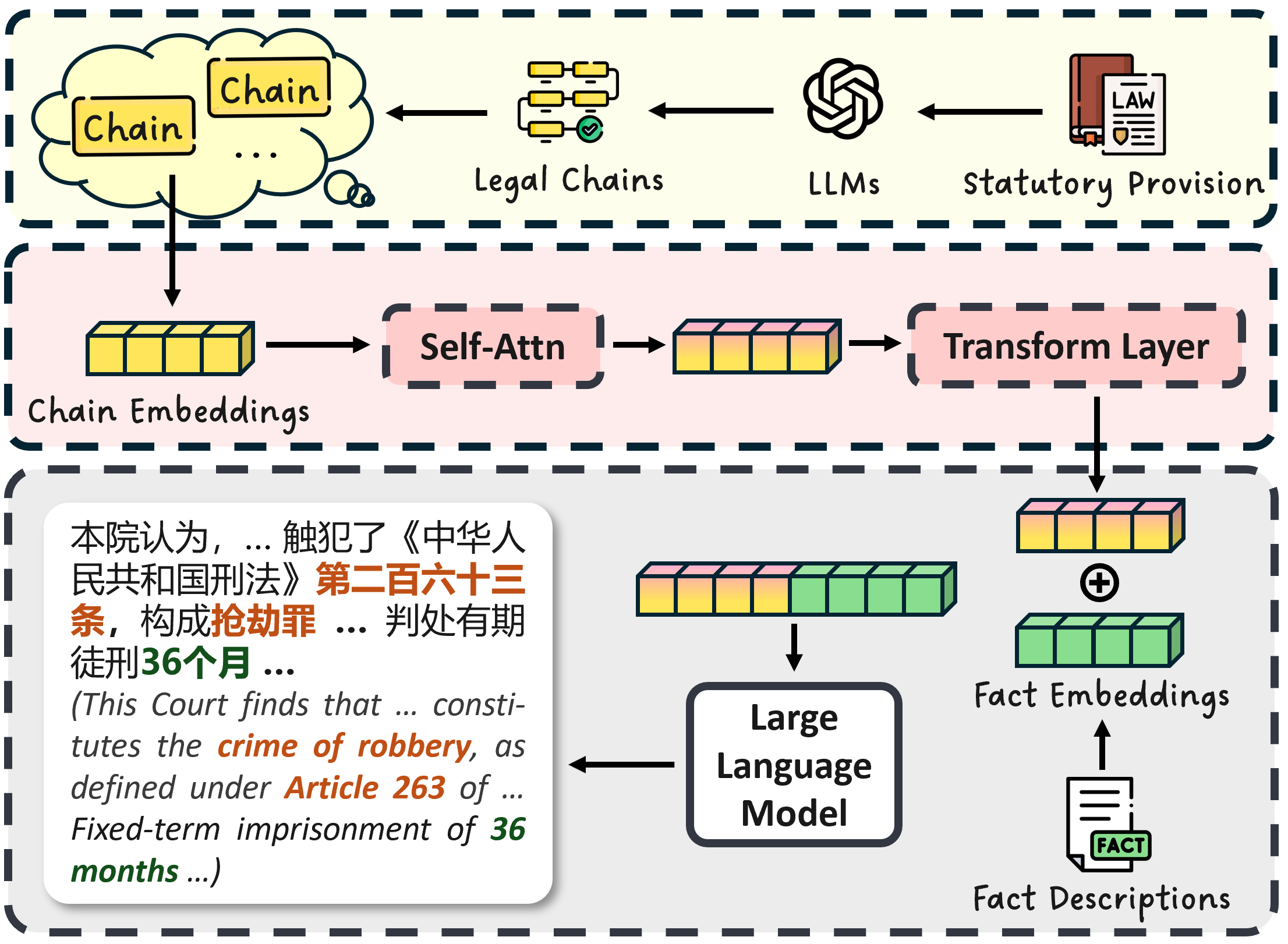}
\caption{The overall architecture of LegalChainReasoner. In the first module (top), legal chains are constructed from statutory provisions. These chains are then processed in the Chain-Aware Encoding module (middle). Finally (bottom), the fused legal knowledge representations are integrated with case fact embeddings through the LLM to generate judicial opinions.}
\vspace{-1em}

\label{fig:legal_chain_structure}
\end{figure}

\paragraph{Overall Framework} The fundamental principle of our method lies in mirroring authentic judicial processes, producing comprehensive judicial opinions that jointly consider both reasoning and sentencing in accordance with established legal principles. Figure~\ref{fig:legal_chain_structure} presents the overall framework of the proposed method, LegalChainReasoner. The framework consists of the Legal Chain construction and the Chain-Aware Encoding modules. The Legal Chain construction addresses the need for formalized legal knowledge representation. By transforming legal provisions into structured \textit{premise-situation-conclusion} triplets, complex legal conditions and their relationships are explicitly formalized, enabling the model to learn step-by-step judicial decision-making processes. The Chain-Aware Encoding module leverages these structured chains to guide the opinion generation process. 
This specialized mechanisms is integrated with case fact embeddings before being passed to the language model. Through this architecture, ensuring the criminal judicial opinion generation process is explicitly grounded in proper legal analysis.


\subsection{Legal Chain}
In criminal law practice, judges typically derive case outcomes through a structured reasoning process. This process traditionally follows a syllogistic pattern~\cite{dai2025laiw}, where \textit{major premises} (statutory provisions) and \textit{minor premises} (factual descriptions) are combined to reach judicial decisions, encompassing both legal reasoning and sentencing. However, judges do not apply statutory provisions in their raw form, as these provisions often contain complex structures, including multiple conditional clauses and nested relationships. To navigate this complexity, legal norm theory~\cite{keuth1974on, raz1980theconcept} offers a principle that decomposes each statutory provision into one or more triplets, each comprising \textbf{a premise, a situation, and a conclusion}. In the context of criminal law, these triplets capture the constitutive elements of a crime and the conditional factors affecting sentencing. Judges typically select and apply the most relevant triplet during the reasoning process to reach a legally sound and context-appropriate judgment.

\begin{figure}[t]
\centering
\includegraphics[width=\linewidth]{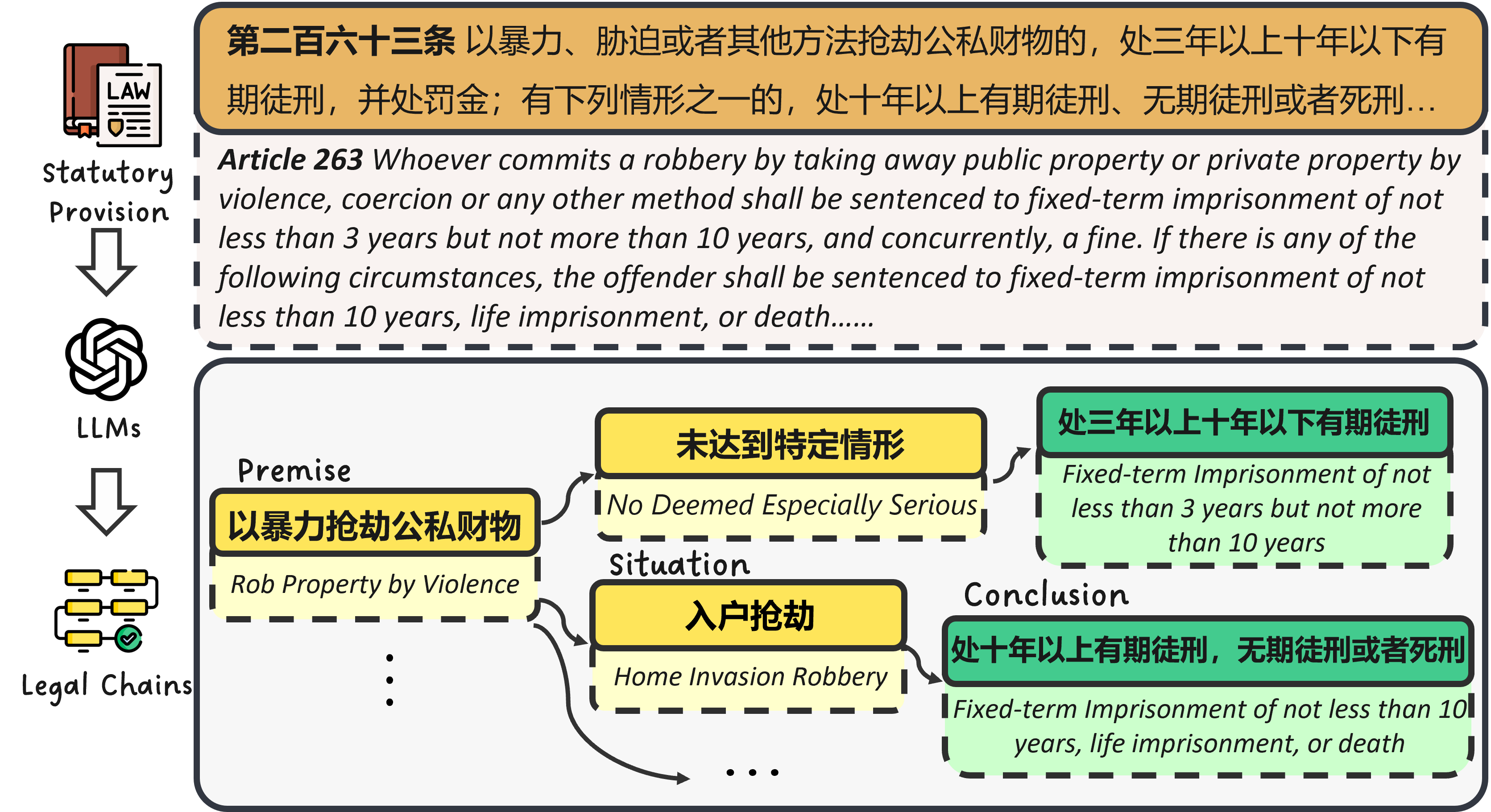}
\caption{Legal Chain construction process with the robbery provisions. The original statutory provision (top) is decomposed into structured premise-situation-conclusion triplets (bottom). 
}
\vspace{-1em}

\label{fig:legal_chain_construction}
\end{figure}

To align with real-world judicial practice, we propose using the triplet structure, \textit{premise}-\textit{situation}-\textit{conclusion}, as a means of integrating legal knowledge into language models for generating judicial opinions. We refer to this triplet as a \textbf{Legal Chain}, which aims to formalize the implicit reasoning patterns embedded within statutory provisions into explicit, structured representations, enabling models to learn the step-by-step judicial decision-making process. 
Figure~\ref{fig:legal_chain_construction} illustrates an example of the legal chains for {\tt robbery}.

Formally, for each criminal charge $C$, all relevant statutory provisions defined within the legal domain are denoted as $\mathcal{SP}=\{SP_1, SP_2, \cdots, SP_j, \cdots\}$. 
They are systematically decomposed into a set of legal chains $\Phi_C = \{\phi_1, \phi_2, \cdots \}$, where each legal chain $\phi\in \Phi_C$ is represented as a triplet $\langle p, s, c \rangle$. Here \(p\) (premise) represents a condition derived from legal facts, typically characterizing a legal element or prerequisite that must be satisfied; \(s\) (situation) captures the consequences and severity assessments, reflecting nuanced legal requirements where multiple factors must be considered simultaneously; and \(c\) (conclusion) specifies a sentencing recommendation directly grounded in statutory provisions.

Current works utilize various forms of legal knowledge, such as legal graphs~\cite{zhong2020jec, yang2021legalgnn}, prediction processes~\cite{xu2024divide}, and explanations of statutory provisions~\cite{yue2021neurjudge}. However, most of these approaches rely on manually designed structures or expert annotations~\cite{chalkidis2019neural}, which significantly increase the cost of incorporating legal knowledge and may result in unstable model performance when the knowledge is incorrect or of low quality. To address this, we propose an automatic extraction process for Legal Chains. This process transforms statutory provisions into structured triplets through a prompt-based extraction methodology using large language models. The extraction prompts are carefully designed with constraints to ensure that the resulting chains maintain both legal accuracy and structural consistency. The complete prompt template and its constraints are provided in Appendix~\ref{sec:appendix_A}. The generated Legal Chains are subsequently validated by legal experts to ensure their fidelity to the original statutory provisions.

Then, for a factual description $F_i$ and its corresponding criminal charge $C_i$, a set of legal chains $\Phi_{C_i} = \{\phi_1, \phi_2, \ldots, \phi_n\}$ is constructed, where each $\phi_j$ represents a distinct decision-making pathway. We did not focus on charge determination, because recent works demonstrate significant proficiency in determining applicable criminal charges, with accuracy rates more than 95\% in criminal charge determination~\cite{hu2018few,zhao2022charge,le2020learning}. Based on this established capability, we focus on more challenging task of generating cohesive judicial opinions that jointly consider both reasoning and sentencing.  When applied to the Criminal Judicial Opinion Generation task, the criminal judicial opinion generation function can then be expressed as $f_{\text{CJOG}}(F_i, C_i, \Phi_{C_i})$.


\subsection{Chain-Aware Encoding}

To better make models understand legal chains $\Phi_{C_i}$, we propose a Chain-Aware Encoding Module to leverage these chains for the Criminal Judicial Opinion Generation task. 


For a given criminal charge $C_i$, each chain $\phi_j = \langle p_j, s_j, c_j \rangle$ in the set $\Phi_{C_i}$ is embedded by first tokenizing and then averaging the token embeddings for each component. 
\begin{equation}
\scalebox{0.80}{$
\mathbf{e}_{p_j} = \frac{1}{|p_j|}\sum_{t \in p_j}\mathbf{E}_{\text{base}}(t), 
$}
\end{equation}
\begin{equation}
\scalebox{0.80}{$
\mathbf{e}_{s_j} = \frac{1}{|s_j|}\sum_{t \in s_j}\mathbf{E}_{\text{base}}(t), 
$}
\end{equation}
\begin{equation}
\scalebox{0.80}{$
\mathbf{e}_{c_j} = \frac{1}{|c_j|}\sum_{t \in c_j}\mathbf{E}_{\text{base}}(t),
$}
\end{equation}
\noindent where, $\mathbf{E}_{\text{base}}$ represents the token embedding function of language models. These component embeddings are then concatenated to form a chain representation $\mathbf{H}_{\phi_j} = [\mathbf{e}_{p_j}; \mathbf{e}_{s_j}; \mathbf{e}_{c_j}] \in \mathbb{R}^{3 \times d}$, where $d$ is the model's hidden dimension.

To capture interactions between the elements of each legal chain, we apply self-attention mechanism:
\begin{equation}
\scalebox{0.80}{$
\mathbf{A}_{\phi_j}, \mathbf{w}_{\phi_j} = \text{MultiHeadAttn}(\mathbf{H}_{\phi_j}), 
$}
\end{equation}
\begin{equation}
\scalebox{0.80}{$
\text{MultiHeadAttn}(\mathbf{H}^h_{\phi_j}) = \text{Atten}(\mathbf{H}^h_{\phi_j}, \mathbf{H}^h_{\phi_j}, \mathbf{H}^h_{\phi_j}),
$}
\end{equation}
\noindent where $\mathbf{A}_{\phi_j} \in \mathbb{R}^{3 \times d}$ is the attended output representation, $\mathbf{w}_{\phi_j} \in \mathbb{R}^{3 \times 3}$ represents the attention weights between chain components, and $h$ means heads of multi-head Attention.

A residual connection preserves both the original and attention-weighted information: $\mathbf{A'}_{\phi_j} = \mathbf{H}_{\phi_j} + \mathbf{A}_{\phi_j}$. This attended representation is then pooled to obtain a single vector representation $\mathbf{r}_{\phi_j} \in \mathbb{R}^{d}$ for each chain. 

After obtaining the processed chain representation $\mathbf{r}_{\phi_j}$, as different types of criminal charges require distinct reasoning patterns based on their statutory provisions and jurisprudential traditions. Inspired by the Mixture-of-Experts (MoE)~\cite{masoudnia2014mixture, cai2024survey}, we design a crime-specific transformation layer to capture specific information. In addition, we introduce a general transformation layer to extract shared, generalizable features. We refer to this module as the crime transformation block, which processes the chain representations. 
The general transformation first applies to all criminal charges:
\begin{equation}
\scalebox{0.80}{$
\mathbf{T}_{\text{G}}(\mathbf{r}_{\phi_j}) = \mathbf{W}_{\text{G}_2} \cdot \text{ReLU}(\mathbf{W}_{\text{G}_1} \cdot \mathbf{r}_{\phi_j} + \mathbf{b}_{\text{G}_1}) + \mathbf{b}_{\text{G}_2}, 
$}
\end{equation}
\noindent where $\mathbf{W}_{\text{G}_1}, \mathbf{W}_{\text{G}_2} \in \mathbb{R}^{d \times d}$ are weight matrices and $\mathbf{b}_{\text{G}_1}, \mathbf{b}_{\text{G}_2} \in \mathbb{R}^{d}$ are bias vectors. Then, for each criminal charge $C_i$, a dedicated transformation is applied to the generally transformed representation:
\begin{equation}
\scalebox{0.80}{$
\mathbf{T}_{C_i}(\mathbf{T}_{\text{G}}(\mathbf{r}_{\phi_j})) = \mathbf{W}_{C_i} \cdot \mathbf{T}_{\text{G}}(\mathbf{r}_{\phi_j}) + \mathbf{b}_{C_i}, 
$}
\end{equation}
\noindent where $\mathbf{W}_{C_i} \in \mathbb{R}^{d \times d}$ and $\mathbf{b}_{C_i} \in \mathbb{R}^{d}$ are charge-specific parameters that encode reasoning patterns unique to each crime type.

These transformations are then combined through a gating mechanism to determine the appropriate balance between general and crime-specific reasoning:
\begin{equation}
\scalebox{0.80}{$
\mathbf{G}(\mathbf{T}_{\text{G}}(\mathbf{r}_{\phi_j})) = \sigma(\mathbf{W}_{\text{g}} \cdot \mathbf{T}_{\text{G}}(\mathbf{r}_{\phi_j}) + \mathbf{b}_{\text{g}}),
$}
\end{equation}
\begin{equation}
\scalebox{0.80}{$
\begin{aligned}
\mathbf{t}_{\phi_j} = {} & \mathbf{G}(\mathbf{T}_{\text{G}}(\mathbf{r}_{\phi_j})) \odot \mathbf{T}_{C_i}(\mathbf{T}_{\text{G}}(\mathbf{r}_{\phi_j})) \\
& + (1 - \mathbf{G}(\mathbf{T}_{\text{G}}(\mathbf{r}_{\phi_j}))) \odot \mathbf{T}_{\text{G}}(\mathbf{r}_{\phi_j}), 
\end{aligned}
$}
\end{equation}
\noindent where $\sigma$ is the sigmoid function that produces content-dependent interpolation weights.

After applying transformations and the gating mechanism, each chain representation is processed through a final fusion layer that combines the original and transformed representations:
\begin{equation}
\scalebox{0.80}{$
\mathbf{f}_{\phi_j} = \mathbf{W}_{\text{fusion}} \cdot [\mathbf{r}_{\phi_j}; \mathbf{t}_{\phi_j}] + \mathbf{b}_{\text{fusion}},
$}
\end{equation}
\noindent where $\mathbf{W}_{\text{fusion}} \in \mathbb{R}^{d \times 2d}$ and $\mathbf{b}_{\text{fusion}} \in \mathbb{R}^{d}$ are learnable parameters.

The final chain representation for all chains is then constructed:
\begin{equation}
\scalebox{0.80}{$
\mathbf{E}_{\text{Chain}} = [\mathbf{f}_{\phi_1}; \mathbf{f}_{\phi_2}; ...; \mathbf{f}_{\phi_n}] \in \mathbb{R}^{n \times d},
$}
\end{equation}
\noindent where $n$ is the number of chains in $\Phi_{C_i}$.

\subsection{Judicial Opinion Generation}

The factual description $F_i$ is embedded using the same embedding layer of language model, and then integrated with the chain representations:
\begin{equation}
\scalebox{0.80}{$
\mathbf{E}_{\text{Combined}} = [\mathbf{E}_{\text{Chain}}; \mathbf{H}_{F_i}] \in \mathbb{R}^{(n+l_{F_i}) \times d},
$}
\end{equation}
\noindent where $\mathbf{H}_{F_i} \in \mathbb{R}^{l_{F_i} \times d}$ represents the embedded tokens of $F_i$ and $l_{F_i}$ is the sequence length. This integrated representation is processed by the language model to generate the judicial opinion $\hat{O}$:
\begin{equation}
\scalebox{0.80}{$
\hat{O} = f_{\text{CJOG}}(F_i, C_i, \Phi_{C_i}) = \text{LLM}(\mathbf{E}_{\text{Combined}}).
$}
\end{equation}

The entire framework is optimized to jointly consider both sentencing prediction accuracy and reasoning quality. The loss function is defined as:
{\small
\begin{equation}
\begin{split}
    \mathcal{L}_{\text{total}} &= \alpha \mathcal{L}_{\text{Reasoning}} + \beta \mathcal{L}_{\text{Sentencing}},\\
    \mathcal{L}_{\text{Reasoning}} &= -\frac{1}{N} \sum_{i=1}^N \sum_{c=1}^C O_{i,c} \log(\hat{O}_{i,c}),\\    \mathcal{L}_{\text{Sentencing}} &= -\frac{1}{N} \sum_{i=1}^N \sum_{c=1}^C S_{i,c} \log(\hat{S}_{i,c}),\\
\end{split}
\end{equation}
}

\noindent where $S$ and $\hat{S}$ are the sentencing part of the judicial opinion. $C$ represent the number of words, and \( \alpha \), \( \beta \) are weighting parameters.

\section{Experiments}

\begin{table*}[t]
\centering
\caption{Performance comparison on the LAIC-2021 dataset. 
GPTScore represents the GPTScore-pairwise. And the GPTScore-pairwise scores for AttS2S and BART reflect their average performance across all comparisons. 
}
\scalebox{0.65}{
\begin{tabular}{lcccccccccc}
\toprule
\textbf{Model} & \textbf{MAE} & \textbf{RMSE} & \textbf{ROUGE-1} & \textbf{ROUGE-2} & \textbf{ROUGE-L} & \textbf{BLEU-1} & \textbf{BLEU-2} & \textbf{BLEU-N} & \textbf{BERTScore} & \textbf{GPTScore} \\
\midrule
AttS2S & 20.86 & 30.96 & 42.93 & 25.24 & 39.04 & 28.25 & 20.34 & 19.12 & 72.72 & 0.00 \\
BART & 28.99 & 41.68 & 51.73 & 35.95 & 46.72 & 39.99 & 32.93 & 31.35 & 79.68 & 20.16 \\
\midrule
Llama-3.2-3B & 17.29 & 28.22 & 45.38 & 27.59 & 40.39 & 33.05 & 25.31 & 23.81 & 78.14 & 4.10 \\
+ Statutory provisions  & 18.88 & 30.99 & 50.56 & 32.25 & 45.05 & 35.52 & 28.28 & 26.75 & 79.63 & 14.90 \\
+ LegalChainReasoner   & \textbf{15.41} & \textbf{25.33} & \textbf{52.82} & \textbf{34.91 }& \textbf{48.00} & \textbf{37.50} & \textbf{30.42} & \textbf{28.85} & \textbf{80.42} & \textbf{52.05} \\
\hline
DeepSeek-R1-Distill-Qwen-7B & 22.47 & 32.06 & 47.64 & 29.44 & 42.61 & 35.91 & 27.90 & 26.16 & 78.87 & 6.61 \\
+ Statutory provisions  & 18.01 & 28.92 & 50.34 & 31.68 & 44.71 & 37.52 & 29.52 & 27.87 & 79.26 & 20.75 \\
+ LegalChainReasoner   & \textbf{15.74} & \textbf{26.17} & \textbf{53.39} & \textbf{34.59} & \textbf{47.71} & \textbf{42.35} & \textbf{33.77} & \textbf{31.94} & \textbf{80.48} & \textbf{50.96} \\
\hline
Lawyer-Llama-13B-V2 & 16.98 & 27.14 & 52.53 & 34.30 & 47.93 & 40.05 & 32.07 & 30.06 & 80.53 & 13.97 \\
+ Statutory provisions  & 14.73 & 25.11 & 54.96 & 36.07 & 49.95 & 42.30 & 34.03 & 32.03 & 81.19 & 25.27 \\
+ LegalChainReasoner   & \textbf{13.54} & \textbf{24.05} & \textbf{55.99} & \textbf{37.28} & \textbf{50.65} & \textbf{44.31} & \textbf{35.89} & \textbf{33.89} & \textbf{81.35} & \textbf{50.88} \\
\bottomrule
\end{tabular}
\label{tab:laic}
}
\end{table*}
\subsection{Experimental Setup}

\paragraph{Datasets.}
In the legal AI domain, there is only one publicly accessible dataset that includes both judicial reasoning and final decisions. It is constructed based on  Legal AI Challenge 2021 (LAIC-2021), which contains Chinese criminal cases featuring factual descriptions, court reasoning, and sentencing outcomes. To ensure a fair comparison during training, we select crime types with more than 500 cases, resulting in approximately 12 distinct crime categories. The detailed statistics of the datasets are provided in Appendix~\ref{app:data}.
To further assess the effectiveness of our proposed approach, we collect a more complex and up-to-date set of cases of the same crime types from the People’s Court Case Database\footnote{Cases on the People's Court Case Database (\url{https://rmfyalk.court.gov.cn}) are selected and published by the Supreme People's Court of China. These cases are intended to guide lower courts in adjudicating similar cases.}, denoted as PCCD. The new dataset includes 100 representative cases (balanced across the 12 crime types) and is used for zero-shot validation, meaning it serves solely to evaluate the models trained on the LAIC-2021.

\paragraph{Baselines.} 

Considering recent studies， we select two types of baselines for a fair comparison. First, we evaluate several natural language generation models, including AttS2S~\cite{bahdanau2015neural}, BART~\cite{lewis2020bart}, Llama-3.2-3B~\cite{grattafiori2024llama3}, DeepSeek-R1-Distill-Qwen-7B~\cite{deepseekai2025deepseekr1}, and Lawyer-Llama-13B-V2~\cite{huang2023lawyer}, on the complete judicial opinion generation task. In addition, we include several widely used and representative methods in sentencing prediction as baselines, such as BERT~\cite{devlin2019bert}, fine-tuned BERT (BERT-FT), Electra~\cite{clark2020electra}, R-Former~\cite{dong2021legal}, NeurJudge~\cite{yue2021neurjudge}, and ML-LJP~\cite{liu2023mlljp}. Further details are provided in Appendix~\ref{app:beselines}.

\paragraph{Evaluation metrics.}
Since the Criminal Judicial Opinion Generation task primarily involves generating text, it can be evaluated using standard and widely adopted text generation metrics, such as ROUGE-1/2/L~\cite{lin2004rouge}, BLEU-1/2/N~\cite{papineni2002bleu}, BERTScore~\cite{zhang2020bertscore}, and GPTScore-pairwise~\cite{li2024gptscore}. Additionally, as sentencing prediction involves numerical outputs, we employ two commonly used regression evaluation metrics: Mean Absolute Error (MAE) and Root Mean Square Error (RMSE). Further details are provided in Appendix~\ref{app:metrics}.

\paragraph{Implementation.}
We apply our LegalChainReasoner to Llama-3.2-3B, DeepSeek-R1-Dstill-Qwen-7B and LawyerLlama-13B-V2. We utilize the Low-Rank Adaptation (LoRA)~\cite{hu2022lora} to fine-tune these LLMs. The setting of LoRA is the same as the previous LLM study in LegalAI~\cite{xu2024divide}. We set 8 heads for the multi-head attention mechanism and 0.1 dropout. As the $\mathcal{L}_{\text{Reasoning}}$ already include the sentence part, to enhance the sentencing part, we found $\alpha=\beta=1$ is the best. We apply Adam optimizer with learning rates at $10^{-5}$ level. All the experiments were conducted on NVIDIA A100 GPUs with 80GB memory. 

\begin{table*}[t]
\centering
\caption{Zero-shot evaluation results on the PCCD dataset. 
GPTScore represents the GPTScore-pairwise. And the GPTScore-pairwise scores for AttS2S and BART reflect their average performance across all comparisons. 
}
\scalebox{0.65}{
\begin{tabular}{lcccccccccc}
\toprule
\textbf{Model} & \textbf{MAE} & \textbf{RMSE} & \textbf{ROUGE-1} & \textbf{ROUGE-2} & \textbf{ROUGE-L} & \textbf{BLEU-1} & \textbf{BLEU-2} & \textbf{BLEU-N} & \textbf{BERTScore} & \textbf{GPTScore} \\
\midrule
AttS2S & 39.13 & 52.96 & 27.20 & 12.54 & 22.85 & 19.28 & 12.17 & 11.88 & 67.02 & 1.33 \\
BART & 62.53 & 77.33 & 29.54 & 15.84 & 23.51 & 24.12 & 16.45 & 15.86 & 70.10 & 25.00 \\
\midrule
Llama-3.2-3B & 38.85 & 52.13 & 28.45 & 12.89 & 21.93 & 21.16 & 13.28 & 12.69 & 68.50 & 1.00 \\
+ Statutory provisions  & 26.53 & 37.24 & 32.76 & 16.13 & 25.88 & 24.04 & 16.22 & 15.50 & 70.17 & 31.00 \\
+ LegalChainReasoner   & \textbf{24.96} & \textbf{36.59} & \textbf{32.96} & \textbf{16.56} & \textbf{26.23} & \textbf{24.25} & 16.39 & 15.70 & \textbf{70.18} & \textbf{32.00} \\
\hline
DeepSeek-R1-Distill-Qwen-7B & 40.71 & 56.41 & 28.13 & 11.91 & 20.60 & 21.75 & 13.06 & 12.47 & 68.11 & 8.00 \\
+ Statutory provisions  & 37.08 & 50.12 & 31.26 & 13.71 & 22.44 & 25.97 & 16.01 & 15.34 & 69.37 & 17.00 \\
+ LegalChainReasoner   & \textbf{33.43} & \textbf{46.52} & \textbf{33.95} & \textbf{16.45} & \textbf{25.68} & \textbf{29.09} & \textbf{18.99} & \textbf{18.19} & \textbf{70.24} & \textbf{45.00} \\
\hline
Lawyer-Llama-13B-V2 & 30.57 & 43.84 & 31.34 & 14.97 & 24.40 & 26.25 & 16.86 & 15.79 & 69.35 & 17.00 \\
+ Statutory provisions  & 27.51 & 39.50 & 33.33 & 16.43 & 26.01 & 26.67 & 17.61 & 16.57 & 69.95 & 27.00 \\
+ LegalChainReasoner   & \textbf{24.96} & \textbf{38.03} & \textbf{33.90} & \textbf{16.84} & \textbf{26.03} & \textbf{27.71} & \textbf{19.39 }& \textbf{17.40} & 70.07 & \textbf{43.00} \\
\bottomrule
\end{tabular}
\label{tab:pccd}
}

\end{table*}

\subsection{Comparison among Criminal Judicial Opinion Generation}
Tables~\ref{tab:laic} and~\ref{tab:pccd} present the performance comparisons between the proposed method and baseline methods on the LAIC-2021 and PCCD datasets for the Criminal Judicial Opinion Generation task.

The proposed LegalChainReasoner, when fused with LLMs, demonstrates superior performance not only in text generation but also in sentence prediction. Across both datasets, the Lawyer Llama variants consistently outperform others, likely due to their pretraining on legal-specific corpora. A performance gap is observed between LAIC-2021 and PCCD across all methods, characterized by higher prediction errors and lower generation quality scores. There are three reasons: (1) PCCD contains more complex cases, as it is published to guide judges; (2) it includes a higher proportion of appellate and final ruling cases involving multiple legal considerations; and (3) these cases need more complex legal reasoning.

When comparing different types of legal knowledge incorporated into LLMs, our proposed legal chain fusion method outperforms models using only statutory provisions or vanilla LLMs. Notably, while fine-tuning a legal-domain LLM can be beneficial, it is not always efficient. By integrating our legal chain approach with general-purpose LLMs such as Llama and DeepSeek, we achieve better performance than the entire fine-tuned Lawyer Llama.

Overall, these results demonstrate the robustness of the proposed framework across diverse legal reasoning scenarios and highlight its effectiveness in formalizing judicial reasoning through premise-situation-conclusion triplets.

\begin{table}[t]
\centering
\caption{Comparison on sentencing prediction. }
\scalebox{0.80}{
\begin{tabular}{lcc}
\hline
\textbf{Method} & \textbf{MAE} & \textbf{RMSE} \\
\hline
NeurJudge & 17.37 & 25.57 \\
R-Former & 23.99 & 32.06 \\
ML-LJP & 20.53 & 30.85 \\
Electra & 23.90 & 32.02 \\
Bert & 21.75 & 31.25 \\
Bert-FT & 23.12 & 31.70 \\
\hline
Llama-3.2-3B & 17.29 & 28.22 \\
+ Statutory provisions  & 18.88 & 30.99 \\
+ LegalChainReasoner   & \textbf{15.41} & \textbf{25.33}\\
\hline
DeepSeek-R1-Distill-Qwen-7B & 22.47 & 32.06 \\
+ Statutory provisions  & 18.01 & 28.92\\
+ LegalChainReasoner   & \textbf{15.97} & \textbf{26.80} \\
\hline
Lawyer-Llama-13B-V2 & 16.98 & 27.14 \\
+ Statutory provisions  & 14.73 & 25.11\\
+ LegalChainReasoner   & \textbf{13.54} & \textbf{24.05} \\
\bottomrule
\end{tabular}
\label{tab:sentencing_prediction_comparison}
}
\vspace{-1em}
\end{table}

\begin{table*}[t]
\centering
\caption{Ablation Study for LegalChainReasoner in Criminal Judicial Opinion Generation. GPTScore represents the GPTScore-pairwise.}
\scalebox{0.65}{
\begin{tabular}{lcccccccccc}
\toprule
\textbf{Model} & \textbf{MAE} & \textbf{RMSE} & \textbf{ROUGE-1} & \textbf{ROUGE-2} & \textbf{ROUGE-L} & \textbf{BLEU-1} & \textbf{BLEU-2} & \textbf{BLEU-N} & \textbf{BERTScore} & \textbf{GPTScore} \\
\midrule
Llama-3.2-3B & 17.29 & 28.22 & 45.38 & 27.59 & 40.39 & 33.05 & 25.31 & 23.81 & 78.14 & 8.79 \\
+ Statutory provisions  & 18.88 & 30.99 & 50.56 & 32.25 & 45.05 & 35.52 & 28.28 & 26.75 & 79.63 & 17.49 \\
+ LegalChain & 16.96 & 29.01 & 48.14 & 29.66 & 42.44 & 34.49 & 26.74 & 25.27 & 78.53 & 9.46 \\
+ LegalChainReasoner   & \textbf{15.41} & \textbf{25.33} & \textbf{52.82} & \textbf{34.91} & \textbf{48.00} & \textbf{37.50} & \textbf{30.42} & \textbf{28.85} & \textbf{80.42} & \textbf{64.26} \\
\bottomrule
\end{tabular}
\label{tab:ablation_study}
}
\vspace{-0.5em}
\end{table*}

\subsection{Comparison among Legal Sentencing Prediction}


Table~\ref{tab:sentencing_prediction_comparison} presents a comparison between the outputs of the proposed method—based on the Criminal Judicial Opinion generation task—and baseline models focused solely on sentencing prediction.

The results indicate that methods based on the Criminal Judicial Opinion generation task generally achieve better average performance. This suggests that treating sentencing prediction in isolation is limited in effectiveness. The proposed LegalChainReasoner framework, when integrated with LLMs, achieves the best overall performance in terms of both MAE and RMSE. This performance gap highlights the importance of incorporating structured legal knowledge for accurate sentencing prediction. Moreover, it demonstrates that our approach not only provides good quality legal reasoning but also outperforms models specifically designed for sentencing prediction.

Among the sentencing prediction models, NeurJudge~\cite{yue2021neurjudge} demonstrated the strongest performance, followed by ML-LJP~\cite{liu2023mlljp}, while the remaining models performed considerably worse. These results differ from those reported in the original papers. Since our formulation treats sentencing prediction as a regression task rather than a classification task, using discrete ranges to represent numerical outcomes may obscure finer-grained differences in model performance and limit the ability to fully assess each model's predictive capability.

It is noteworthy that the simple incorporation of statutory provisions without structured reasoning chains produced inconsistent results. In some cases, such as with Llama-3.2-3B~\cite{grattafiori2024llama3}, direct addition of statutory provisions actually degraded sentencing accuracy. This observation underscores the insufficiency of merely exposing models to legal text without proper structuring of the reasoning process.

\subsection{Ablation Study}

Table~\ref{tab:ablation_study} shows the comparison for different components in the proposed method, LegalChainReasoner, in Criminal Judicial Opinion Generation task based on the Llama-3.2-3B.

Overall, only using all components can achieve better results. Incorporating statutory provisions alone created "legal knowledge overload," improving text generation but degrading sentencing prediction due to lack of structured guidance. When fusing legal chains as plain text, more balanced performance was achieved as the premise-situation-conclusion structure provided effective reasoning pathways. When using the proposed legal chain-aware encoding method, performance can be further enhanced by capturing interdependencies between legal elements. 


\subsection{Case Study}

\begin{table}[t]
\centering
\caption{Case Study Evaluation Results for Criminal Judicial Opinion Generation}
\scalebox{0.60}{
\begin{tabular}{lccccc}
\hline
\textbf{Model} & \textbf{Defendant} & \textbf{Situation} & \textbf{Sentencing} & \textbf{Combined} \\
 & \textbf{Accuracy} & \textbf{Accuracy} & \textbf{Accuracy} & \textbf{Performance} \\
 & \textbf{(\%)} & \textbf{(\%)} & \textbf{(\%)} & \textbf{Score (\%)} \\
\hline
AttS2S & 8.45 & 42.26 & 76.15 & 2.72 \\
BART & 99.50 & 65.27 & 12.22 & 7.93 \\
Lawyer-Llama-13B-V2 & 99.08 & 56.82 & 71.30 & 40.12 \\
+ Statutory provisions & \textbf{99.41} & 63.18 & 74.39 & 46.66 \\
+ LegalChainReasoner & \textbf{99.41} & \textbf{67.20} & \textbf{78.49} & \textbf{52.39} \\
\hline
\end{tabular}
\label{tab:case_study_results}
}
\end{table}

\paragraph{Rule-based Screening.} To evaluate the quality of generated judicial opinions, we design a rule-based screening method that assesses the content across three critical dimensions: defendant identification accuracy, situation description correctness, and sentencing range appropriateness. More details are provided in Appendix~\ref{app:rule}. Table~\ref{tab:case_study_results} shows that while most models achieve high accuracy in defendant identification, significant differences occur in situation description and sentencing accuracy. LegalChainReasoner demonstrates better performance in situation description while maintaining strong accuracy in sentencing, resulting in the highest overall combined performance score.

\begin{figure}[t]
\centering
\scalebox{0.66}{
\renewcommand{\arraystretch}{1.3}
\setlength{\tabcolsep}{4pt}
\small
\begin{tabular}{|p{1.4cm}|p{1.3cm}|p{2.5cm}|p{3cm}|}
\hline
\textbf{Model} & \textbf{Defendant} & \textbf{Reasoning} & \textbf{Sentencing} \\
\hline
\textbf{AttS2S} & 
\cellcolor{MorandiRed}\textbf{Incorrect} \newline Incorrectly identify & 
\cellcolor{MorandiRed}\textbf{Misclassified} \newline Fraud instead of robbery & 
\cellcolor{MorandiYellow}\textbf{Inconsistent} \newline Sentencing inconsistent with reasoning \\
\hline
\textbf{BART} & 
\cellcolor{MorandiGreen}\textbf{Correct} \newline Correctly identify & 
\cellcolor{MorandiGreen}\textbf{Correct} \newline Used violence and threats to rob & 
\cellcolor{MorandiRed}\textbf{Inconsistent} \newline Sentencing inconsistent with reasoning and actual sentencing range \\
\hline
\textbf{Lawyer-Llama-13B-V2} & 
\cellcolor{MorandiGreen}\textbf{Correct} \newline Correctly identify & 
\cellcolor{MorandiGreen}\textbf{Correct} \newline Used violence and threats to rob & 
\cellcolor{MorandiYellow}\textbf{Consistent} \newline Sentencing consistent with reasoning \\
\hline
\textbf{+ Statutory Provisions} & 
\cellcolor{MorandiGreen}\textbf{Correct} \newline Correctly identify & 
\cellcolor{MorandiGreen}\textbf{Correct} \newline Used violence and threats to rob & 
\cellcolor{MorandiRed}\textbf{Inconsistent} \newline Sentencing inconsistent with reasoning and actual sentencing range \\
\hline
\textbf{+ LegalChainReasoner} & 
\cellcolor{MorandiGreen}\textbf{Correct} \newline Correctly identify & 
\cellcolor{MorandiGreen}\textbf{Correct} \newline Used violence and threats to rob & 
\cellcolor{MorandiGreen}\textbf{Consistent} \newline Sentencing consistent with reasoning and most close \\
\hline
\multicolumn{4}{|p{9cm}|}
{\footnotesize 
\textbf{Case Facts:} The court ascertained that on January 2, 2012, at approximately 22:30, defendant Fu XX, together with Cui Zhongyu and Ren Zhenguo (both already sentenced), drove a black Honda sedan to the Shell gas station in Tangerlizhen, Bazhou City. Wearing masks and armed with knives and pickaxe handles, they threatened [staff] and robbed over 300 yuan in cash from the station. At approximately 1:00 on January 3, the aforementioned individuals drove to the Shell gas station on East Ring Road in Bazhou City and used the same methods to rob over 1,000 yuan in cash. The stolen money has been spent.
}\\
\hline
\end{tabular}
}
\caption{Case Study Concise Example of Generated Criminal Judicial Opinions.}
\label{tab:case_study_example}
\end{figure}

\paragraph{Case Demonstration.} Figure~\ref{tab:case_study_example} and Appendix~\ref{sec:appendix_B} present several concise comparisons of outputs from different models in three types of crimes. Figure~\ref{tab:case_study_example} only shows the summarization of the generated output; the full output is shown in Table~\ref{app:exp1}.
From these cases, we find the AttS2S generated opinion completely misclassifies the crime as fraud rather than robbery, incorrectly identifies the defendant as "Li XX," and fails to acknowledge key elements such as the use of violence or threats, despite recommending a sentence (36 months) that happens to be relatively close to the actual sentence (42 months). The BART~\cite{lewis2020bart} generated opinion correctly identifies the defendant and crime type but recommends a significantly lower sentence (30 months) than the statutory minimum (36 months), as Chinese Criminal Law Article 263 explicitly stipulates that robbery offenses shall be punished by "fixed-term imprisonment of not less than three years and not more than ten years." This legal inconsistency demonstrates the critical limitations of approaches lacking legal reasoning for ensuring sentencing determinations comply with basic statutory requirements. In contrast, the LegalChainReasoner generated opinion demonstrates comprehensive coherence across all dimensions. It correctly identifies the defendant and crime type, acknowledges the use of violence and threats, and recommends a sentence (48 months) that, while slightly higher than the actual sentence, remains proportionate to the case severity and logically aligned with the presented reasoning. 

These findings show that the proposed method enables the model to maintain consistent alignment between fact analysis, legal reasoning, and sentencing determination.


\section{Related Work}

\paragraph{Legal Reasoning Generation} Several studies have formalized legal reasoning as a text-to-text natural language generation task, proposing various approaches to enhance performance. Early work applys sequence-to-sequence models to generate legal reasoning~\cite{ye2018interpretable}. Subsequent research leveraged pre-trained language models such as BART~\cite{lewis2020bart}, including versions fine-tuned on legal datasets~\cite{yue2021circumstances}, as well as large language models~\cite{xu2024divide, yue2024event}. Beyond language model applications, recent studies have incorporated different forms of legal knowledge, either implicitly or through explicit processing. This includes multi-step reasoning~\cite{wu2022towards, xu2024divide} and legal provisions understanding~\cite{li2024enhancing, yue2024event}. However, integrating preprocessed legal knowledge presents several challenges. First, reliance on human experts for annotation and validation incurs high costs, limiting scalability. Then, preprocessing steps may introduce biases or oversimplifications, reducing the robustness of generated reasoning, especially using the LLMs.

\paragraph{Legal Sentencing Prediciton} Legal sentencing prediction in criminal domains targets predicting prison terms. Several types of models are used in this task, including word-to-embedding methods like the Word2Vec-based approach combined with legal corpora~\cite{xia2019similarity}, RNN-variants~\cite{xu2020distinguish}, language models~\cite{zhang2023contrastive,sun2024chinese} and LLMs~\cite{deng2024enabling}. The knowledge used in this area can also be divided into explicit and processed categories. The processed knowledge includes explanatory statutory provision~\cite{yue2021neurjudge} and statute embeddings~\cite{meng2025dpfsi}, similar cases retrieved through Case Triple Modeling~\cite{zhang2023contrastive}.

\section{Conclusion}
This paper introduces the Criminal Judicial Opinion Generation task to bridge the gap between legal reasoning and sentencing in current works by aligning them with actual judicial workflows. To address this and flexibly fuse legal knowledge, the LegalChainReasoner framework is proposed, which models statutory provisions as structured legal chains and applies a chain-aware encoding method to capture legal logic and charge-specific reasoning. Experiments on two real-world and open-source Chinese criminal cases show that LegalChainReasoner improves both reasoning quality and sentencing accuracy, whether the complete judicial opinion generation task or a single task. This work marks a step forward in building models that generate judicial opinions with stronger legal coherence and real-world applicability.


\section*{Limitations}
While LegalChainReasoner demonstrates significant improvements in criminal judicial opinion generation, there are two limitations in evaluation methodology and temporal considerations.





While our automated evaluation using LLMs provides scalability, it may not capture all nuances of legal reasoning that human judges would consider. The combined accuracy metric, though comprehensive, treats all three components (defendant identification, situation description, and sentencing range) as equally important, which may not reflect actual judicial priorities.

The dataset used in this study reflects legal practices and interpretations from a specific time period. Legal frameworks evolve over time through legislative amendments and changing judicial interpretations. The model's ability to adapt to such temporal shifts in legal understanding without retraining remains unexplored.

\section*{Ethic Statements}
\textbf{Privacy Statement} All cases used in this paper are publicly accessible and have been published through China Judgment Online or the People's Court Case Database. All personally identifiable information has been anonymized, as demonstrated in the case studies.

\textbf{Usage Statement} This work is intended solely for research purposes. We have no intention of applying the current version in the real court to judge or influence people. While our proposed method outperforms baseline models, it still contains errors and requires further improvement.



\bibliography{custom}

\appendix

\section{Legal Chain Extraction Prompt}
\label{sec:appendix_A}

\begin{figure}[ht]
    \centering
    \includegraphics[width=0.99\linewidth]{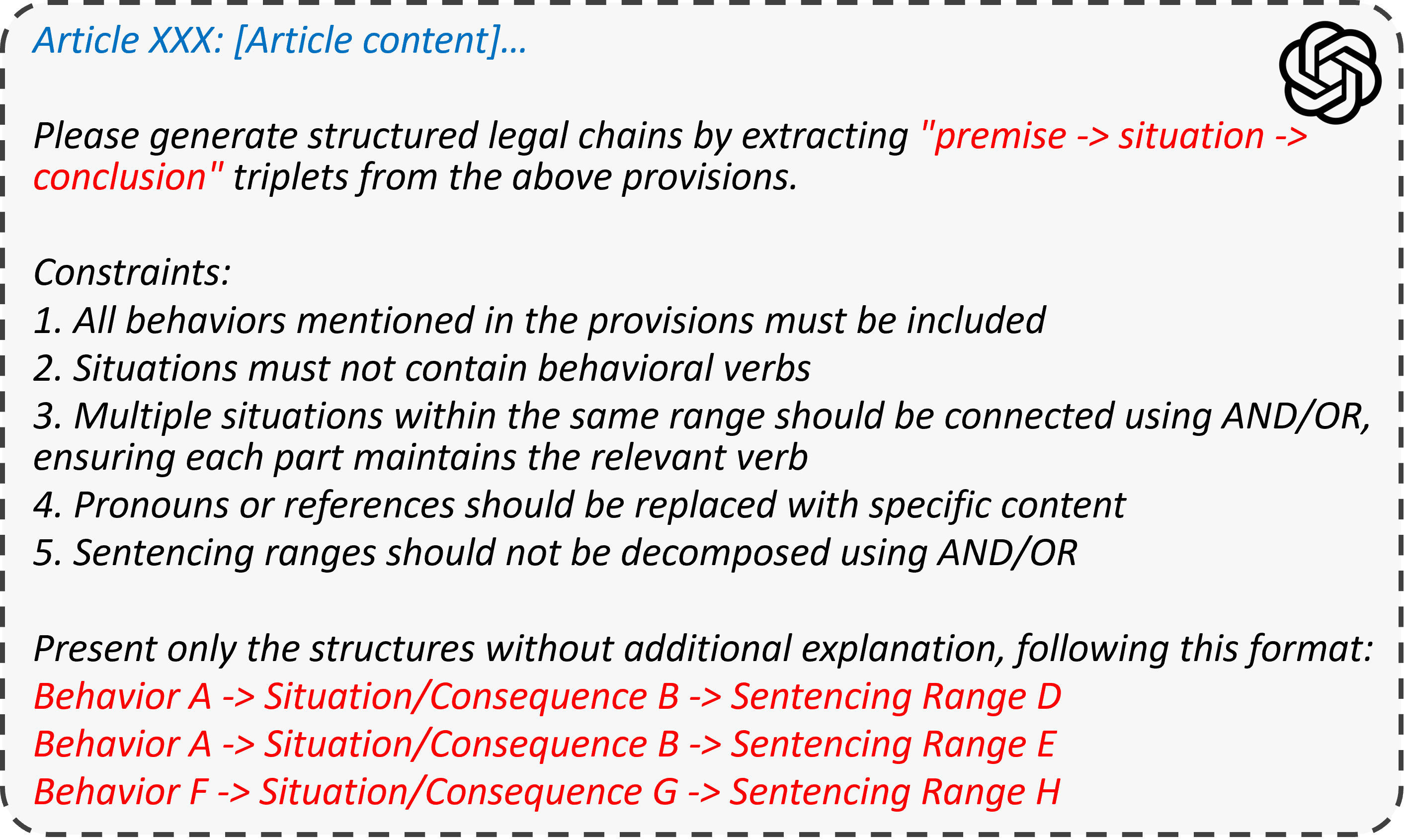}
    \caption{Legal Chain Extraction Prompt.}
    \label{fig:chainprompt}
\end{figure}

Figure~\ref{fig:chainprompt} illustrates the prompt structure used for extracting legal chains from statutory provisions. While the extraction prompts are designed to enforce five essential constraints:
\begin{itemize}
    \item \textbf{Exhaustiveness}: All distinct behaviors and conditions specified in the statutory provision must be represented in the extracted chains.
    \item \textbf{Semantic separation}: Premises must capture behavioral elements, while situations must focus exclusively on consequential states or circumstances.
    \item \textbf{Logical coherence}: Complex conditions are represented through explicit logical operators (AND/OR), ensuring preservation of the original provision's inferential structure. 
    \item \textbf{Referential specificity}: Any pronouns or indirect references in the statutory text must be replaced with their specific content to eliminate ambiguity.
    \item \textbf{Sentencing specificity}: Conclusions must directly correspond to legally prescribed sentencing ranges without logical decomposition.
\end{itemize}















We asked legal experts to evaluate the generated legal chains using the prompt shown in Figure~\ref{fig:chainprompt}. Across all 12 crime categories, the generated legal chains achieved an accuracy of 97.68\%. This high accuracy indicates that the proposed prompt effectively generates accurate legal chains. Moreover, our construction method allows legal knowledge to be represented more easily and systematically than in previous studies, which relied on manually designed rules.

\section{Datasets} \label{app:data}

We utilize two open-source, real-world datasets: the Legal AI Challenge 2021 (LAIC-2021) and the People’s Court Case Database (PCCD). Both datasets contain cases published by courts in China. The key difference is that the PCCD dataset was curated by the Supreme People’s Court of China in 2023 and is intended to guide judges in handling similar cases. Therefore, the cases of PCCD are much more complex, and the number of cases is much smaller than LAIC-2021. Statistics for the two datasets are shown in Table~\ref{tab:dataset_stats}.

\begin{table}[t]
\centering
\small
\caption{Statistics of the Datasets}
\label{tab:dataset_stats}
\begin{tabular}{lccc}
\toprule
& \textbf{LAIC-Train} & \textbf{LAIC-Test} & \textbf{PCCD} \\
\midrule
\textbf{Total Cases} & 33,841 & 1,200 & 100 \\
\midrule
\multicolumn{4}{l}{\textbf{Text Length (Characters)}} \\
\textbf{Fact Description} & & & \\
Average & 548.6 & 509.6 & 394.0 \\
Median & 321.0 & 306.5 & 321.5 \\
\textbf{Court Opinion} & & & \\
Average & 343.6 & 366.1 & 394.0 \\
Median & 280.0 & 299.0 & 332.0 \\
\bottomrule
\end{tabular}
\label{tab:data}
\end{table}

\section{Baelines} \label{app:beselines}

We selected two types of baselines to compare with our method: Criminal Judicial Opinion Generation and Sentencing Prediction. Since the legal reasoning generation task is a text generation task, the models used are primarily language models, which can also be applied to the Criminal Judicial Opinion Generation task.

\textit{Criminal Judicial Opinion Generation task.} Based on prior work in legal reasoning generation, we selected several recent and influential natural language generation models as baselines, including AttS2S~\cite{bahdanau2015neural}, BART~\cite{lewis2020bart}, Llama-3.2-3B~\cite{grattafiori2024llama3}, DeepSeek-R1-Distill-Qwen-7B~\cite{deepseekai2025deepseekr1}, and Lawyer-Llama-13B-V2~\cite{huang2023lawyer}.

\textit{Sentencing Prediction} We selected language models such as BERT~\cite{devlin2019bert}, fine-tuned BERT (BERT-FT), and ELECTRA~\cite{clark2020electra}; manually designed knowledge-based methods like NeurJudge~\cite{yue2021neurjudge} and ML-LJP~\cite{liu2023mlljp}; and case-based methods such as R-Former~\cite{dong2021legal}.

\section{Evaluation Metrics} \label{app:metrics}

We select several types of evaluation metrics to assess the quality of the generated judicial opinions.

\subsection{Numerical Evaluation}
For sentencing prediction, we do not set the task as a classification problem. Although prior work has often formulated sentence length prediction as a classification task by discretizing prison terms into predefined categories, we argue that this approach introduces several limitations. First, discretization can result in the loss of fine-grained information, especially when sentence lengths vary continuously across a wide range. This may obscure meaningful differences between cases with similar but not identical sentence durations. Second, the boundaries between categories are often arbitrary and may not reflect real-world sentencing practices or legal standards, potentially introducing bias or reducing interpretability. Finally, treating the task as regression allows for a more precise, data-driven estimation of sentence length, which aligns better with the inherently continuous nature of the sentencing variable. Therefore, we adopt a regression-based formulation and evaluate performance using Mean Absolute Error (MAE) and Root Mean Square Error (RMSE).

\subsection{Lexical Evaluation}
To assess the quality of legal reasoning generation from a lexical perspective, we utilize standard n-gram overlap-based metrics. Specifically, we use ROUGE-1, ROUGE-2, and ROUGE-L to measure unigram, bigram, and longest common subsequence overlaps, respectively. Additionally, we report BLEU-1, BLEU-2, and BLEU-N scores to evaluate n-gram precision at different levels. These metrics capture surface-level similarity between the generated and reference texts.

\subsection{Embedding Evaluation}
To assess semantic similarity beyond surface-level lexical overlap, we utilize BERTScore~\cite{zhang2020berscore}, which leverages contextualized embeddings from pre-trained transformer models to compare generated and reference judicial opinions. This metric captures deeper semantic alignment and is particularly effective in identifying meaning-preserving variations that traditional n-gram-based metrics may overlook.
 
\subsection{LLM-based Evaluation}
\begin{figure}
    \centering
    \includegraphics[width=1\linewidth]{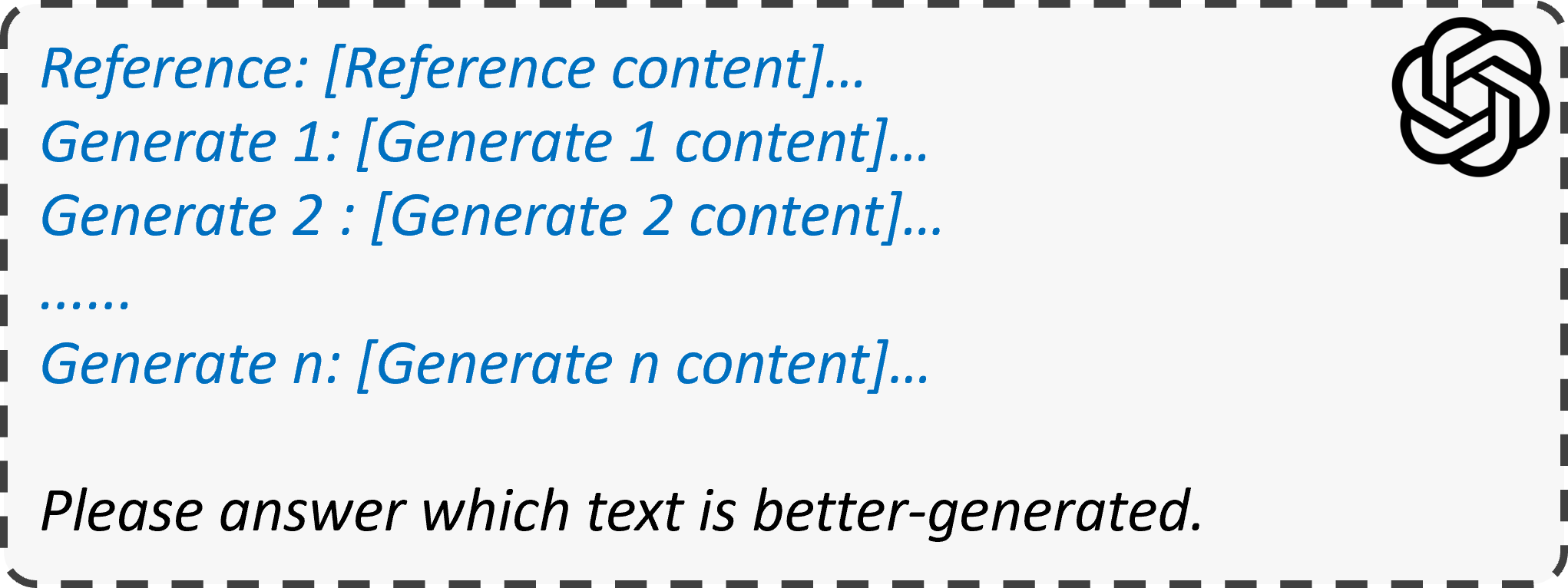}
    \caption{GPTScore-pairwise Prompt.}
    \label{fig:gptscore}
\end{figure}
To evaluate the logical consistency and legal soundness of the generated content, we adopt an LLM-based assessment method known as GPTScore~\cite{fu2024gptscore}. This approach leverages large language models to assess the overall quality of generated judicial opinions. Following recent research, we employ the GPTScore-pairwise~\cite{li2024gptscore} variant, which has been shown to be more effective. This method compares multiple generated texts and selects the best output. Figure~\ref{fig:gptscore} illustrates the prompt used for pairwise comparison, based on prior work.

Table~\ref{tab:laic} and~\ref{tab:pccd} show the scores of each model on GPTScore-pairwise. Due to the page limit, we report the average scores of AttS2S and BART. The AttS2S and BART scores of Table~\ref{tab:laic} are 0.0/0.0/0.0 and 9.87/21.68/28.97, respectively. The AttS2S and BART scores of Table~\ref{tab:pccd} are 2.0/1.0/1.0 and 11/29/35, respectively. 

\subsection{Rule-based Evaluation} \label{app:rule}

We design a rule-based method to evaluate the information consistency between the reference and generated judicial opinions.

This method focuses on three key aspects: defendant, situation, and sentencing. First, we compare the defendant's name (e.g., "Li XX") to ensure it is consistently represented in both the reference and generated opinions. Second, we assess whether both opinions describe the same situation, based on structured legal chains extracted from the statutory provisions. Finally, given the situation mentioned in the generated opinion, we verify whether the predicted sentence aligns with the expected sentencing level for that situation.

The final consistency score is computed by multiplying the individual accuracies of the defendant, situation, and sentencing components.

\section{Case Study}
\label{sec:appendix_B}

\onecolumn
\small
\begin{longtable}[t]{p{1.5cm}|p{13.5cm}}
\caption{Case Demonstration Example 1\label{app:exp1}}\\
\hline
\textbf{Component} & \textbf{Content} \\
\hline
\multicolumn{2}{l}{\textbf{Case Information}} \\
\hline
Type & Robbery \\
\hline
Facts & 经审理查明，2012年1月2日22时30分许，被告人A伙同B、C（二人已判刑）驾驶一辆黑色轿车至X市Y镇1号加油站，蒙面持刀、工具，并以言语相威胁，抢走该站现金300余元。1月3日凌晨1时许，上述几人又驾车至X市Z路2号加油站，以同样手段抢走现金1000余元。赃款已挥霍。

The court ascertained that on January 2, 2012, at approximately 22:30, Defendant A, together with B and C (both already sentenced), drove a black sedan to Gas Station 1 in Town Y, City X. Wearing masks and armed with knives and tools, they threatened the staff and robbed over 300 yuan in cash from the station. At approximately 1:00 on January 3, the aforementioned individuals drove to Gas Station 2 on Road Z in City X and used the same methods to rob over 1,000 yuan in cash. The stolen money has been spent. \\
\hline
Opinion & 本院认为，被告人A以非法占有为目的，以暴力威胁手段抢劫他人财物，其行为已构成抢劫罪。公诉机关指控的罪名成立，本院依法予以支持。被告人A持械抢劫，酌情从重处罚。被告人A当庭自愿认罪，酌情从轻处罚。依照《中华人民共和国刑法》第二百六十三条，第二十五条第一款，第五十二条，第五十三条之规定，判决如下：判处有期徒刑42个月。

This court holds that Defendant A, with the purpose of illegal possession, used violence and threats to rob others' property. His actions constitute the crime of robbery. The prosecution's charge is established and is supported by this court. Defendant A committed armed robbery, which warrants a heavier punishment. Defendant A voluntarily admitted guilt in court, which warrants a lenient punishment. In accordance with Articles 263, 25(1), 52, and 53 of the Criminal Law of the People's Republic of China, the judgment is as follows: 42 months of fixed-term imprisonment. \\
\hline
\multicolumn{2}{l}{\textbf{Evaluation Metrics}} \\
\hline
Defendant & Defendant A \\
\hline
Reasoning & Robbery (with knives and tools); Multiple instances (two gas stations)
\\
\hline
Sentencing & 42 months of imprisonment \\
\hline
\multicolumn{2}{l}{\textbf{Generated Opinions}} \\
\hline
AttS2S & \textbf{Opinion:} This court holds that \colorbox{MorandiRed}{Defendant D}, with the purpose of illegal possession, used methods of fabricating facts and concealing the truth to \colorbox{MorandiRed}{defraud others' property}, \colorbox{MorandiRed}{with the amount being huge}. These actions \colorbox{MorandiRed}{constitute the crime of fraud}. The prosecution's charge is established. The defendant truthfully confessed his crimes, which warrants a lenient punishment. Based on the defendant's criminal facts, circumstances, and remorseful attitude, a lenient punishment is appropriate. In accordance with Articles 266, 67(1), and 62 of the Criminal Law of the People's Republic of China, the judgment is as follows: \colorbox{MorandiYellow}{36 months of fixed-term imprisonment}.

\textbf{Evaluation:} \scalebox{0.85}[1]{$\times$}~Incorrectly identifies defendant as "Defendant D"; 
\scalebox{0.85}[1]{$\times$}~Completely mischaracterizes the crime as fraud and fails to mention used violence and threats as premise; 
\scalebox{0.85}[1]{$\times$}~Sentencing inconsistent with reasoning \\
\hline

BART & \textbf{Opinion:} This court holds that \colorbox{MorandiGreen}{Defendant A}, with the purpose of illegal possession, together with others \colorbox{MorandiGreen}{used violence and threats} to \colorbox{MorandiGreen}{forcibly rob others' property}. These actions \colorbox{MorandiGreen}{constitute the crime of robbery}. The prosecution's charge is established. The Defendant A played a minor role in the joint crime and is an accomplice who should receive reduced punishment according to law. The Defendant A truthfully confessed his crimes after being arrested, which warrants a lenient punishment according to law. In accordance with Articles 263, 25(1), 27, 67(3), and 64 of the Criminal Law of the People's Republic of China, the judgment is as follows: \colorbox{MorandiRed}{30 months of fixed-term imprisonment}.

\textbf{Evaluation:} \checkmark~Correctly identifies defendant; \checkmark~Correctly identifies crime as robbery and mention used violence and threats as premise; \scalebox{0.85}[1]{$\times$}~Sentencing inconsistent with reasoning and actual sentencing range \\
\hline

Lawyer-Llama-13B-V2 & \textbf{Opinion:} This court holds that \colorbox{MorandiGreen}{Defendant A}, together with others, with the purpose of illegal possession, \colorbox{MorandiGreen}{used violence and threats} to \colorbox{MorandiGreen}{rob public and private property}. His actions \colorbox{MorandiGreen}{constitute the crime of robbery}. The prosecution's charge is supported by clear facts and sufficient evidence. The Defendant A voluntarily admitted guilt in court, which warrants consideration for leniency. In accordance with Articles 263(1) and 55(1) of the Criminal Law of the People's Republic of China, the judgment is as follows: \colorbox{MorandiYellow}{60 months of fixed-term imprisonment}.

\textbf{Evaluation:} \checkmark~Correctly identifies defendant; \checkmark~Correctly identifies crime as robbery and mention used violence and threats as premise; \checkmark~Sentencing consistent with reasoning though higher than actual sentence \\
\hline

Lawyer-Llama-13B-V2 + Statutory Provisions & \textbf{Opinion:} This court holds that \colorbox{MorandiGreen}{Defendant A}, together with others, with the purpose of illegal possession, \colorbox{MorandiGreen}{used violence and threats} to \colorbox{MorandiGreen}{rob public and private property}. His actions \colorbox{MorandiGreen}{constitute the crime of robbery}. The prosecution's charge is supported by clear facts and sufficient evidence. The Defendant A voluntarily admitted guilt in court, which warrants consideration for leniency. In accordance with Article 263(4) of the Criminal Law of the People's Republic of China, the judgment is as follows: \colorbox{MorandiRed}{120 months of fixed-term imprisonment}.

\textbf{Evaluation:} \checkmark~Correctly identifies defendant; \checkmark~Correctly identifies crime as robbery and mention used violence and threats as premise; \scalebox{0.85}[1]{$\times$}~Sentencing inconsistent with reasoning and actual sentencing range \\
\hline

Lawyer-Llama-13B-V2 + Legal-Chain-Reasoner & \textbf{Opinion:} This court holds that \colorbox{MorandiGreen}{Defendant A}, together with others, with the purpose of illegal possession, \colorbox{MorandiGreen}{used violence and threats} to \colorbox{MorandiGreen}{rob public and private property}. His actions \colorbox{MorandiGreen}{constitute the crime of robbery}. The prosecution's charge is supported by clear facts and sufficient evidence. The Defendant A voluntarily admitted guilt in court, which warrants consideration for leniency. In accordance with Article 263 of the Criminal Law of the People's Republic of China, the judgment is as follows: \colorbox{MorandiGreen}{48 months of fixed-term imprisonment}.

\textbf{Evaluation:} \checkmark~Correctly identifies defendant; \checkmark~Correctly identifies crime as robbery and mention used violence and threats as premise; \checkmark~Sentencing consistent with reasoning though higher than actual sentence but most closed \\
\hline
\end{longtable}

\vspace{6em}

\begin{longtable}[t]{p{1.5cm}|p{13.5cm}}
\caption{Case Demonstration Example 2}\\
\hline
\textbf{Component} & \textbf{Content} \\
\hline
\multicolumn{2}{l}{\textbf{Case Information}} \\
\hline
Type & Fraud \\
\hline
Facts & 经审理查明，被告人A于2013年1月至4月，以能为被害人B办理工厂上班为名，两次骗取被害人B人民币16万元。被告人A于2014年11月份，以给C的儿子办理工作为由，骗取C人民币24万元。被告人A于2017年2月份，以帮助D办事为由先后三次骗取D人民币22万元。综上，被告人A共诈骗三起，诈骗金额计人民币62万元。

The court ascertained that from January to April 2013, Defendant A defrauded Victim B by claiming he could arrange employment at Factory X for Victim B's son, receiving 160,000 CNY on two occasions.
In November 2014, the defendant defrauded Victim C of 240,000 CNY by claiming he could arrange employment for Victim C's son.
In February 2017, the defendant defrauded Victim D of 220,000 CNY on three occasions by claiming he could help with certain matters.
In total, Defendant A committed three instances of fraud amounting to 620,000 CNY. \\
\hline
Opinion & 本院认为，公诉机关指控被告人A以非法占有为目的，虚构事实，隐瞒真相，多次骗取他人财物，数额特别巨大的事实，有经过庭审核实的证据予以证实，A的行为构成诈骗罪，公诉机关指控的犯罪事实和罪名成立。关于A辩解称，其没有承诺为C儿子办工作，收到C24万元系借款，其行为不构成诈骗罪的意见，经审理认为，被害人C、证人E、证人F证言均证实A谎称能为C儿子办工作，C共给付A24万元，书证收据（欠条），载明为出款人孩子办工作；短信截图证实A谎称为C儿子办理工作，让C拿钱的事实，上述证据能够形成完整证据链条，足以证实A谎称能为C儿子办工作，诈骗C24万元的事实，故A的辩解不成立；关于A辩解称，其在D处分三次拿22万元后让G拿走了，钱没花着，不能认定其构成诈骗罪的辩护意见，经审理认为，被害人D的陈述、证人G的证言均证实将22万元交付给A，且G对A称其将22万元借走一事予以否认亦无证据证实，故A的辩解不成立。被告人A诈骗被害人B案系刑满释放后五年内重新犯罪，是累犯，应当从重处罚；被告人A如实供述诈骗被害人B的罪行，可以从轻处罚。依照《中华人民共和国刑法》第二百六十六条、第六十五条第一款、第六十七条第三款、第五十二条、第五十三条、第六十四条之规定，判决如下：判处有期徒刑138个月。

This court holds that the prosecution's charges against Defendant A—that with the purpose of illegal possession, he fabricated facts, concealed the truth, and repeatedly defrauded others of property in an especially large amount—are supported by evidence verified during the trial. Defendant A's actions constitute the crime of fraud, and the prosecution's charges regarding the criminal facts and offense are established.
Regarding Defendant A's defense that he did not promise to arrange employment for Victim C's son and that the 240,000 CNY received from Victim C was a loan rather than fraud, the court finds that testimonies from Victim C, Witness E, and Witness F all confirm that Defendant A falsely claimed he could arrange employment for Victim C's son, and Victim C paid Defendant A 240,000 CNY in total. Documentary evidence including receipts (IOUs) indicate the money was for arranging employment for the payer's child. Text message screenshots confirm Defendant A's false claims about arranging employment and requesting money from Victim C. This evidence forms a complete chain sufficient to prove that Defendant A defrauded Victim C of 240,000 CNY by falsely claiming he could arrange employment for Victim C's son. Therefore, Defendant A's defense is not established.
Regarding Defendant A's defense that after receiving 220,000 CNY from Victim D in three installments, he let Witness G take it away and did not spend the money himself, thereby not constituting fraud, the court finds that testimonies from Victim D and Witness G confirm that 220,000 CNY was given to Defendant A, and Witness G denies Defendant A's claim that he borrowed the 220,000 CNY. There is no evidence to support Defendant A's claim, so his defense is not established.
The defendant committed the fraud against Victim B after being released from prison for less than five years, making him a recidivist, which warrants a heavier punishment. The defendant truthfully confessed to defrauding Victim B, which warrants a lenient punishment. In accordance with Articles 266, 65(1), 67(3), 52, 53, and 64 of the Criminal Law of the People's Republic of China, the judgment is as follows: 138 months of fixed-term imprisonment. \\
\hline
\textbf{Component} & \textbf{Content} \\
\hline
\multicolumn{2}{l}{\textbf{Evaluation Metrics}} \\
\hline
Defendant & Defendant A \\
\hline
Reasoning & Multiple defrauded others of property (three separate victims); Especially large amount (620,000 CNY)
\\
\hline
Sentencing & 138 months of imprisonment \\
\hline
\multicolumn{2}{l}{\textbf{Generated Opinions}} \\
\hline
AttS2S & \textbf{Opinion:} The Court holds that \colorbox{MorandiRed}{Defendant Z}, with the purpose of illegal possession, used methods of fabricating facts and concealing the truth to \colorbox{MorandiGreen}{defraud others' property}, \colorbox{MorandiRed}{with the amount being large}. These actions \colorbox{MorandiGreen}{constitute the crime of fraud}. The prosecution's charge is established. The defendant truthfully confessed his crimes after committing them, which warrants lenient punishment according to law. Based on the defendant's criminal facts, circumstances, and remorseful attitude, a lenient punishment is appropriate. In accordance with Articles 266, 67(1), and 62 of the Criminal Law of the People's Republic of China, the judgment is as follows: \colorbox{MorandiRed}{36 months of fixed-term imprisonment}.

\textbf{Evaluation:} \scalebox{0.85}[1]{$\times$}~Incorrectly identifies defendant; \scalebox{0.85}[1]{$\times$}~Correctly classifies premise as defraud others' property but mischaracterizes amount as "large" rather than "especially large"; \scalebox{0.85}[1]{$\times$}~Sentencing inconsistent with reasoning and actual sentencing range \\
\hline

BART & \textbf{Opinion:} The Court holds that \colorbox{MorandiGreen}{Defendant A}, with the purpose of illegal possession, used methods of fabricating facts and concealing the truth to \colorbox{MorandiGreen}{defraud others' property}, \colorbox{MorandiGreen}{with the amount being especially large}. These actions \colorbox{MorandiGreen}{constitute the crime of fraud}. The prosecution's charge against the defendant for the crime of fraud is supported by clear facts and sufficient evidence. The charge is established and supported by this court. The defendant truthfully confessed his crimes after being arrested, which warrants lenient punishment according to law. In accordance with Articles 266, 67(3), 52, 53, and 64 of the Criminal Law of the People's Republic of China, the judgment is as follows: \colorbox{MorandiYellow}{120 months of fixed-term imprisonment}.

\textbf{Evaluation:} \checkmark~Correctly identifies defendant; \checkmark~Correctly classifies premise as defraud others' property and correctly characterizes amount as "especially large"; \checkmark~Sentencing consistent with reasoning though lower than actual sentence \\
\hline

Lawyer-Llama-13B-V2 & \textbf{Opinion:} The Court holds that \colorbox{MorandiGreen}{Defendant A}, with the purpose of illegal possession, fabricated facts and concealed the truth, to \colorbox{MorandiGreen}{repeatedly defraud others' property}, \colorbox{MorandiGreen}{with the amount being especially large}. These actions \colorbox{MorandiGreen}{constitute the crime of fraud}. The prosecution's charge is supported by evidence and is established. The defendant truthfully confessed his crimes after being arrested, which constitutes a confession and warrants lenient punishment according to law. The defendant has actively returned the defrauded property and obtained forgiveness from the victims, which may warrant lenient punishment. In accordance with Articles 266 and 67(3) of the Criminal Law of the People's Republic of China, the judgment is as follows: \colorbox{MorandiRed}{96 months of fixed-term imprisonment}.

\textbf{Evaluation:} \checkmark~Correctly identifies defendant; \checkmark~Correctly classifies premise as defraud others' property and correctly characterizes amount as "especially large"; \scalebox{0.85}[1]{$\times$}~Sentencing inconsistent with reasoning and actual sentencing range \\
\hline

Lawyer-Llama-13B-V2 + Statutory Provisions & \textbf{Opinion:} The Court holds that \colorbox{MorandiGreen}{Defendant A}, with the purpose of illegal possession, fabricated facts and concealed the truth to \colorbox{MorandiGreen}{defraud others' property}, \colorbox{MorandiGreen}{with the amount being especially large}. These actions \colorbox{MorandiGreen}{constitute the crime of fraud}. The prosecution's charge is supported by clear facts and sufficient evidence, and is established. The defendant voluntarily admitted guilt in court, which warrants lenient punishment. After the case occurred, the defendant's family actively compensated the victims for economic losses and obtained forgiveness, which warrants lenient punishment. The defense counsel's relevant defense opinions are consistent with the established facts and are adopted. In accordance with Articles 266, 52, and 64 of the Criminal Law of the People's Republic of China, the judgment is as follows: \colorbox{MorandiYellow}{120 months of fixed-term imprisonment}.

\textbf{Evaluation:} \checkmark Correctly identifies defendant; \checkmark Correctly classifies premise as defraud others' property and correctly characterizes amount as "especially large"; \checkmark Sentencing consistent with reasoning though lower than actual sentence\\
\hline

Lawyer-Llama-13B-V2 + Legal-Chain-Reasoner & \textbf{Opinion:} The Court holds that \colorbox{MorandiGreen}{Defendant A}, with the purpose of illegal possession, fabricated facts and concealed the truth to \colorbox{MorandiGreen}{defraud others' property}, \colorbox{MorandiGreen}{with the amount being especially large}. These actions \colorbox{MorandiGreen}{constitute the crime of fraud}. The prosecution's charge is supported by clear facts, sufficient evidence, and is established. Given that the defendant voluntarily admitted guilt in court, a relatively lenient punishment may be considered. In accordance with Articles 266, 52, and 53 of the Criminal Law of the People's Republic of China, the judgment is as follows: \colorbox{MorandiYellow}{120 months of fixed-term imprisonment}.

\textbf{Evaluation:} \checkmark~Correctly identifies defendant; \checkmark~Correctly classifies premise as defraud others' property and correctly characterizes amount as "especially large"; \checkmark~Sentencing consistent with reasoning though lower than actual sentence \\
\hline
\end{longtable}

\begin{longtable}[t]{p{1.5cm}|p{13.5cm}}
\caption{Case Demonstration Example 3}\\
\hline
\textbf{Component} & \textbf{Content} \\
\hline
\multicolumn{2}{l}{\textbf{Case Information}} \\
\hline
Type & Misappropriation of Funds \\
\hline
Facts & 经审理查明：2017年3月5日，被告人A与B某签定股权债权转让协议书，在2017年3月10日至3月13日，B某分三次通过银行转账转入被告人A账户150万元的合同应付款，被告人A将150万元挪用还银行贷款。2017年4月21日至5月9日，B某将X公司Y分公司的售房款分6次转入被告人A账户106万元，被告人A将其中85万元售房款用于还自己的贷款和个人使用。案发前已归还。另查明，被害人X公司及X公司Y分公司均表示对被告人A谅解。

Facts as determined by the court: On March 5, 2017, Defendant A signed an equity and debt transfer agreement with Person B. Between March 10-13, 2017, Person B transferred contractual payments totaling 1.5 million yuan to Defendant A's account in three separate bank transfers, which Defendant A misappropriated to repay bank loans. From April 21 to May 9, 2017, Person B transferred housing sales proceeds totaling 1.06 million yuan from Branch Y of Company X to Defendant A's account in six installments. Defendant A used 850,000 yuan of these funds to repay personal loans and for personal expenses. The funds were returned before the case was filed. Further investigation revealed that the victims, Company X and its Branch Y, have both expressed understanding toward Defendant A \\
\hline

Opinion & 本院认为，被告人A利用职务之便挪用资金，数额较大，犯罪事实清楚，证据确实、充分，其行为构成挪用资金罪。公诉机关指控的罪名及事实成立，本院予以确认。被告人A到案后如实供述自己的罪行，可从轻处罚。被告人A获得被害人谅解，可酌情从轻处罚。公诉机关的量刑建议符合法律规定，本院予以确认。辩护人关于被告人如实供述、获得被害人谅解、认罪认罚可从轻处罚的辩护意见予以采纳。辩护人关于起诉书指控第一起被告人挪用150万元不应认定及第二起犯罪数额应为45万元的辩护意见，经查与事实不符，不予采纳。根据被告人犯罪的事实、犯罪的性质、情节和对于社会的危害程度，依照《中华人民共和国刑法》第二百七十二条、第六十四条、第六十七条第三款之规定，判决如下：判处有期徒刑7个月。

This court holds that Defendant A took advantage of his position to misappropriate funds in a relatively large amount. The facts of the crime are clear and the evidence is reliable and sufficient. His behavior constitutes the crime of misappropriation of funds. The prosecution's charge and alleged facts are established and confirmed by this court.
The defendant truthfully confessed his crimes after being arrested, which warrants a lenient punishment. The defendant obtained the forgiveness of the victims, which also warrants consideration for leniency. The sentencing recommendation of the prosecution complies with the law and is confirmed by the court.
The defense's opinion that the defendant truthfully confessed, obtained the victims' understanding, and admitted guilt and accepted punishment warranting leniency is adopted. The defense's opinion regarding the prosecution's allegations about the first misappropriation of 1.5 million yuan and that the amount in the second crime should be 450,000 yuan is not consistent with the facts and is not adopted.
Based on the facts, nature, circumstances, and social harm of the defendant's crime, in accordance with Articles 272, 64, and 67(3) of the Criminal Law of the People's Republic of China, the judgment is as follows: 7 months of fixed-term imprisonment. 
\\
\hline
\textbf{Component} & \textbf{Content} \\
\hline
\multicolumn{2}{l}{\textbf{Evaluation Metrics}} \\
\hline
Defendant & Defendant A \\
\hline
Reasoning & Misappropriated funds; Relatively large amounts
\\
\hline
Sentencing & 7 months of imprisonment \\
\hline
\multicolumn{2}{l}{\textbf{Generated Opinions}} \\
\hline
AttS2S & \textbf{Opinion:} This court holds that \colorbox{MorandiRed}{Defendant Z}, with the purpose of illegal possession, used methods of fabricating facts and concealing the truth to \colorbox{MorandiRed}{defraud others' property}, \colorbox{MorandiRed}{with the amount being large}. These actions \colorbox{MorandiRed}{constitute the crime of fraud}. The prosecution's charge is established. The defendant's criminal facts and circumstances warrant lenient punishment according to law. Based on the defendant's criminal facts, circumstances, and the degree of social harm, in accordance with Articles 266, 67(1), and 60 of the Criminal Law of the People's Republic of China, the judgment is as follows: \colorbox{MorandiRed}{36 months of fixed-term imprisonment}.

\textbf{Evaluation:} \scalebox{0.85}[1]{$\times$}~Incorrectly identifies defendant; \scalebox{0.85}[1]{$\times$}~Mischaracterizes crime as fraud rather than misappropriation of funds and incorrectly characterizes amount as "large" rather than "relatively large"; \scalebox{0.85}[1]{$\times$}Sentencing inconsistent with reasoning and actual sentencing range \\
\hline

BART & \textbf{Opinion:} This court holds that \colorbox{MorandiGreen}{Defendant A}, as a company employee, took advantage of his position to \colorbox{MorandiGreen}{misappropriate company funds} for personal use, \colorbox{MorandiRed}{with the amount being large}, and not returned for over three months. His actions \colorbox{MorandiGreen}{constitute the crime of misappropriation of funds}. The prosecution's charge is established and supported by this court. The defendant truthfully confessed his crimes after being arrested, which warrants lenient punishment according to law. The defendant has obtained the understanding of the victims, which warrants consideration for leniency. In accordance with Articles 272(1), 67(3), and 64 of the Criminal Law of the People's Republic of China, the judgment is as follows: \colorbox{MorandiRed}{36 months of fixed-term imprisonment}.

\textbf{Evaluation:} \checkmark~Correctly identifies defendant; \scalebox{0.85}[1]{$\times$}~Correctly identifies premise as misappropriate company funds but incorrectly characterizes amount as "large" rather than "relatively large"; \scalebox{0.85}[1]{$\times$}~Sentencing inconsistent with reasoning and actual sentencing range\\
\hline

Lawyer-Llama-13B-V2  & \textbf{Opinion:} This court holds that \colorbox{MorandiGreen}{Defendant A} took advantage of his position to \colorbox{MorandiRed}{misappropriate public funds} for profit-making activities, \colorbox{MorandiRed}{with the amount being large}. His actions \colorbox{MorandiGreen}{constitute the crime of misappropriation of funds}. The prosecution's charge is established with clear facts and sufficient evidence. The defendant voluntarily surrendered himself after the crime and truthfully confessed, constituting voluntary surrender, which warrants lenient punishment. The defendant voluntarily admitted guilt and obtained the victims' understanding, which also warrants consideration for leniency. In accordance with Articles 272(1), 67(1), 72(1), 72(3), 73(2), and 73(3) of the Criminal Law of the People's Republic of China, the judgment is as follows: \colorbox{MorandiRed}{36 months of fixed-term imprisonment}.

\textbf{Evaluation:} \checkmark~Correctly identifies defendant; \scalebox{0.85}[1]{$\times$}~Incorrectly characterizes as "misappropriation of public funds" rather than "misappropriation of company funds" and incorrectly describes amount as "large" rather than "relatively large"; \scalebox{0.85}[1]{$\times$}Sentencing inconsistent with reasoning and actual sentencing range \\
\hline

Lawyer-Llama-13B-V2 + Statutory Provisions & \textbf{Opinion:} This court holds that \colorbox{MorandiGreen}{Defendant A} took advantage of his position to \colorbox{MorandiGreen}{misappropriate company funds} for personal use, \colorbox{MorandiGreen}{with the amount being relatively large} and not returned for over three months. His actions \colorbox{MorandiGreen}{constitute the crime of misappropriation of funds}. The prosecution's charge is established with clear facts and sufficient evidence. Before the case was filed, the defendant returned most of the misappropriated funds to the victim company and obtained their understanding, which warrants consideration for leniency. The defendant voluntarily admitted guilt during the trial, which also warrants leniency. The defense counsel's opinion that the defendant is a first-time offender who has actively compensated for the victims' losses and obtained their understanding, requesting a suspended sentence, is verified and adopted. In accordance with Articles 272(1), 67(3), 72(1), 73(2), and 73(3) of the Criminal Law of the People's Republic of China, the judgment is as follows: \colorbox{MorandiYellow}{12 months of fixed-term imprisonment}.

\textbf{Evaluation:} \checkmark~Correctly identifies defendant; \checkmark~Correctly identifies premise as misappropriation of funds and correctly characterizes amount as "relatively large"; \checkmark~Sentencing consistent with reasoning though higher than actual sentence \\
\hline

Lawyer-Llama-13B-V2 + Legal-Chain-Reasoner & \textbf{Opinion:} This court holds that \colorbox{MorandiGreen}{Defendant A} took advantage of his position to \colorbox{MorandiGreen}{misappropriate company funds} for personal use, \colorbox{MorandiGreen}{with the amount being relatively large} and not returned for over three months. His actions \colorbox{MorandiGreen}{constitute the crime of misappropriation of funds}. The prosecution's charge and alleged facts are established and legally supported. Given that the defendant actively returned all misappropriated funds before the case was filed, voluntarily admitted guilt in court, and obtained the understanding of the victims, a lenient punishment is warranted. In accordance with Articles 272(1) and 64 of the Criminal Law of the People's Republic of China, the judgment is as follows: \colorbox{MorandiYellow}{12 months of fixed-term imprisonment}.

\textbf{Evaluation:} \checkmark~Correctly identifies defendant; \checkmark~Correctly identifies premise as misappropriation of funds and correctly characterizes amount as "relatively large"; \checkmark~Sentencing consistent with reasoning though higher than actual sentence \\
\hline
\end{longtable}

\twocolumn
\end{CJK*}
\end{document}